%% file: main.tex
\documentclass[a4paper,twoside]{article}      
\usepackage{amsmath,amssymb,amsfonts,amsthm} 
\usepackage{graphics}                 
\usepackage{color}                    
\usepackage{hyperref}                 
\usepackage{comment} 			
\usepackage{enumitem}

\theoremstyle{plain}
\newtheorem{theorem}{Theorem}[section]

\theoremstyle{definition}
\newtheorem{definition}{Definition}[section]
\newtheorem{conjj}[definition]{Conjecture}

\theoremstyle{definition}
\newtheorem{example}{Example}[section]

\date{\small\it \today}
\title{Algebraic Expression of Subjective Spatial and Temporal Patterns
\footnote{Great thanks for whole heart support of my wife. Thanks for Internet and research contents contributers to Internet.}}

\author{ Chuyu Xiong \\
{\small Independent researcher, New York, USA} \\
{\small Email: chuyux99@gmail.com}
}

\begin{document}
\maketitle
\begin{abstract}
\input abs
\end{abstract}

\bigskip

\section{Introduction}
\input intro
\section{Objective Patterns}\label{table}
\input opattern

\section{Subjective Patterns}
\input spattern

\section{X-forms}
\input xform

\section{Future Works}
\input more

\end{document}

%% file: abs.tex
Universal learning machine is a theory trying to study machine learning from mathematical point of view.  The outside world is reflected inside an universal learning machine according to pattern of incoming data. This is subjective pattern of learning machine. In  \cite{paper2, cpaper}, we discussed subjective spatial pattern, and established a powerful tool -- X-form, which is an algebraic expression for subjective spatial pattern. However, as the initial stage of study, there we only discussed spatial pattern. Here, we will discuss spatial and temporal patterns, and algebraic expression for them. 

%% file: intro.tex
Machine learning, especially deep learning, currently is a very active area of research. However, unfortunately, a thorough mathematical foundation for deep learning is still missing. For example, what deep learning really is doing inside is still quite mystery. Universal learning machine is a mathematical theory to study machine learning by trying to understand patterns. By definition, an universal learning machine is a machine can learn any pattern from data. Pattern can be spatial, and can be temporal. In order to reduce the complexity, we focus on spatial pattern only at first. We developed theory of universal learning machine for only spatial pattern \cite{paper2, cpaper}. 

In the studies in \cite{paper2, cpaper}, we found that one very crucial step is to find a way to express the subjective view of a learning machine to its outside world. Thus, we introduced one effective tool, which we called as X-form. X-form is an algebraic expression of subjective spatial pattern. By using X-form, we indeed get deeper understanding of learning machine. For example, we can define the data sufficiency for learning, which tell us what kind of data are sufficient for learning. Another example, we achieve to understand what deep learning is really doing \cite{paper3}. Based on X-form, we were able to introduce some effective learning strategies and methods, and we proved that universal learning machine indeed exists. 

Now, taking the restriction on spatial pattern away, we are going to discuss universal learning machine for spatial and temporal pattern, i.e. any pattern. Following the same approach we did before, we first focus on the very crucial part: algebraic expression of subjective spatial and temporal pattern.

First, let us explain universal learning machine. See illustration of universal learning machine in Fig. 1. 

As in the illustration, universal learning machine has major components: input space, output space, conceiving space and governing space. 

\begin{center}
\begin{picture}(300,230)(0,0)
\put(0,0){\resizebox{11 cm}{!}{\includegraphics{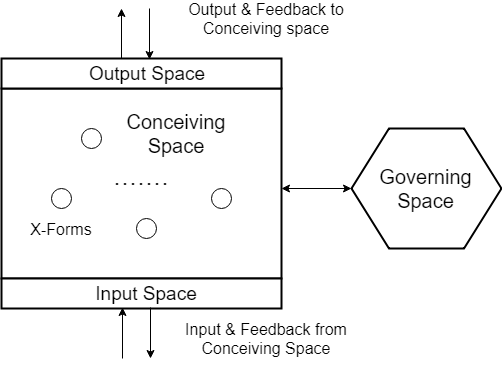}}}
\end{picture}

{\bf Fig. 1. Illustration of Universal Learning Machine} 
\end{center}

Input space is a N-dim binary space. So, any input is an N-dim binary array. But, since we are talking about spatial and temporal, we should study sequence of input. We do not restrict length of sequence. For effective learning machine, input space also gives feedback from learning machine to its input enviornment. But, we do not need to consider this feedback in this article.

Output space is one M-dim binary space. There could be some feedback to learning machine for its output. In this article, we do not consider this feedback either.

Between input space and output space lays conceiving space. This is the major part of learning machine, information processing from input to output is done here, and learning is done here. Conceiving space has the capability to handle spatial and temporal patterns, and is able to make spatial and temporal memory, and take spatial and temporal abstraction. Conceiving space is actually the container of X-forms.

Governing space is where rules of learning stay. Governing space is very crucial. 

Following, first we study objective patterns, spatial and temporal. For temporal pattern, understanding of sequence is crucial. We will introduce notation for sequence. Then we turn to subjective: subjective spatial pattern, subjective temporal pattern and subjective pattern. Based on these discussions, we reach algebraic expression for subjetive patterns, sX-form, tX-form and X-form. We prove the central theorem: any spatial and temporal pattern (literally, any pattern) can be expression by X-form established on some very limited base patterns.




%% file: opattern.tex
Inside a learning machine there is information processing unit that takes input and generates output. So it takes information from input and generates output accordingly. But, how to take information? Must according to pattern of input. But, what is pattern, spatial pattern, temporal pattern? 

Consider a learning machine $M$ and one input $p$. $p$ is N-dim binary array. So, $p$ has components cross the array. Such data arrangement cross over the array is called spatial. Clearly, the spatial arrangement contains information, and $M$ could process according to such information. If $M$ treats $p$ only by the information carried by spatial data arrangement, not by any other,  such as previous input, afterward input, input order, etc, it is called spatial. If $M$ treats $p$ also according to other factor, such as input order, previous input, afterward input, etc, it is called temporal. It is very clear that spatial pattern involves just the spatial data arrangement in input array, while temporal pattern must involve spatial data arrangement, and temporal factors.
 
So, if we talk about input $p$ for spatial pattern, only need to consider $p$ itself. If we talk about input of a temporal pattern, $p$ is only a part. We have to consider more. For temporal pattern, we need to consider sequence, such as: $\{p_1, p_2, \ldots, p_k\}$. 

\begin{example}[\bf Simple Sequence] Some very simple examples. \\
$[0 \to 1 \to 0 \to 0 \to 1]$ is one simple sequence. Here the input space is very simple, i.e. 1-dim binary space.  The length of sequence is 5.\\
$[(0,0) \to (0,1) \to (1,0) \to (0,0)]$ is another very simple sequence, here the input space is 2-dim binary space. The length of sequence is 4. 
\end{example} 
Note, we use such symbols to express sequence: $[ \cdots]$ is to denote one sequence, and inside $[ ]$ are data, i.e. a series of input, and $\to$ to denote the relationship of "next". In sequence, the relationship "NEXT" is crucial. We will use these symbols for sequence.

\subsection{Base Pattern, Sequence and Pattern}

Spatial and temporal pattern is classified by just one input or sequence of input. But, pattern can also be classified by objective or subjective. Objective means that patterns are independent from any particular learning machine. In order to describe it well, we start from base pattern space.

\begin{definition}[\bf Base Pattern Space]
$N$-dim base pattern space, denote as $PS^0_N$,  is a $N$-dim binary vector space, i.e.  
$$
PS^0_N = \{b = (i_1, i_2, \ldots, i_N) \ | \  i_k = 0 \ or \ 1 \}
$$ 
\end{definition}

Each element of $b \in PS^0_N$ is a base pattern. There are totally $2^N$ base patterns in $PS^0_N$. When $N$ is not very small, $PS^0_N$ is a huge set. For example, if $N = 100$, which is very small comparing to need of any machine, then $2^N = 2^{100} > 10^{33}$. This is a huge number.

Base pattern is just the starting point. Pattern is much more than base pattern. However, since an objective spatial pattern is totally independent from any particular machine, it could not be anything but a set of base patterns. So, we have:

\begin{definition}[\bf Objective Spatial Pattern as Set of Base Patterns]
An objective spatial pattern $p$ is a set of base patterns: 
$$
p = \{ b_1, b_2, \ldots | b_i \in  PS^0_N \}
$$
Any member of the set $p$, $b \in p$, is called an instance of $p$. 
\end{definition}
Note, before we used term "face of a pattern". We now use term "instance", which should be better. 

Before defining objective temporal pattern, we need to consider sequence of base patterns, so we introduce some notations for sequence of base patterns. We do not restrict the length of sequence. Notation $PS^0_N*$ stands for unspecified folds of $PS^0_N$. As usual, symbol $*$ means unspecified folds. 

\begin{definition}[\bf Base Pattern Column]
$PS^0_N*$ is called as base pattern column. It is unspecified folds of base pattern space. More precisely, any member $s \in PS^0_N*$ is a sequence of base patterns, written as: 
$$
s = [b_1 \to b_2 \to \ldots \to b_k], \ \ b_j \in PS^0_N, \  j = 1, \ldots, k
$$
for some integer $k$. 
\end{definition}
Any member in base pattern column is a sequence of base patterns, whose length is a finite number $k$, but this number $k$ could be any. Also, in above definition, we use one notation for a sequence:
$$
s = [b_1 \to b_2 \to \ldots \to b_k]
$$
That is to say, the symbol $[\cdots]$ stands for a sequence, and inside $[\ \ ]$, we put contents of this sequence, i.e. base patterns connected by symbol "$\to$". We call the symbol "$\to$" as {\it next}. 

Parallel to objective spatial pattern, we have objective temporal pattern.
 
\begin{definition}[\bf Objective Temporal Pattern as Set of Sequence of Base Patterns]
An objective temporal pattern $p$ is a set of sequence of base patterns: 
$$
p = \{ s_1, s_2, \ldots | s_i \in  PS^0_N* \}
$$
Any member of $p$, $s \in p$ is called an instance of $p$. 
\end{definition}
Any member of the set $p$, $s \in p$, is a sequence like this: $s = [b_1 \to b_2 \to \cdots \to b_k], \ b_j \in  PS^0_N, \  j = 1, \ldots, k$. Such a sequence is an instance of $p$. Also note, for 2 members of $p$, both of them are sequence. But, their lengths of sequence are no need to be same.

One very special case of temporal pattern is the pattern that has length 1. Actually, we can just thought it as a spatial pattern. We can also think any spatial pattern as a temporal pattern with length 1, and any base pattern as a sequence with length 1.

\begin{definition}[\bf Objective Spatial Pattern as Objective Temporal Pattern]
A objective spatial pattern $p$ is a set of base patterns, $p = \{ b_1, b_2, \ldots | b_i \in  PS^0_N \}$. However, $p$ can also be viewed as a temporal pattern that each sequence has length 1, i.e. only has one spatial pattern: $tp = [p] = \{ [b_1], [b_2], \ldots \}$.
\end{definition}
We call such temporal pattern as length 1 temporal pattern. 

\begin{definition}[\bf Objective Pattern (Spatial and Temporal)]
A objective pattern (spatial and temporal) $p$ is such a set:  $p = \{ p_1, p_2, \ldots \}$, where $p_j$ is one instance of $p$, $p_j$ is sequence of base patterns (sequence length of $p_j$ could be 1, in that case, $p_j$ is just a base pattern). 
\end{definition}

This definition tells us that objective pattern $p$ is just a subset of base pattern column, each instance of $p$ is just a sequence of base patterns (if the length is 1, just base pattern).  

Now we turn to operators of objective pattern.

\subsection{Operators}
For objective spatial patterns, we have defined operators 3 operators: OR, AND, NOT in \cite{paper2, cpaper}. We now modify these operators so that they are good for any objective pattern (spatial and temporal). 

\begin{definition}[\bf Operator OR (set union)]
For any 2 objective patterns $p_1$ and $p_2$ , we have a new pattern $p$: 
$$
p = p_1 \ OR \ p_2 = p_{1_b} \cup p_{2_b} 
$$
Here, $\cup$ is the set union. 
\end{definition}
That is to say, this new pattern is such a pattern that any instance it is either instance of $p_1$ or instance of $p_2$.  

\begin{definition}[\bf Operator AND (set intersection)]
For any 2 objective patterns $p_1$ and $p_2$ , we define a new pattern $p$: 
$$
p = p_1 \ AND \ p_2 = p_{1_b} \cap p_{2_b} 
$$
Here, $\cap$ is the set intersection. 
\end{definition}
That is to say, this new pattern is such a pattern that any instance of it is both instance of $p_1$ and instance of $p_2$.  

\begin{definition}[\bf Operator NOT (set complement) ]
For any objective patterns $p$ , we define a new pattern $p$: 
$$
q = NOT \ p = p_{b} ^c = \{ b \in PS^0_N* | b \not \in p_{b} \}  
$$
Here, $A^c$ is complement set of $A$. Also note, the complement set is taken in base pattern column $PS^0_N*$.
\end{definition}
That is to say, $q$ is such a pattern that its instance is not an instance of $p$.  

Since objective patterns are subset of base pattern column, it is no surprise to realize that these operators are actually set operators: set union, set intersection, and set complement. All set operators are in base pattern column.

Very clearly, the above 3 operators do not depend on any learning machine. So, they are all objective. Consequently, if we apply these 3 operators consecutively, we still generate a new objective pattern. 

But, we have another important operator for objective patterns (spatial and temporal). 

\begin{definition}[\bf Operator NEXT ]
For any 2 objective patterns $p_1$ and $p_2$, we define a new pattern $p$: 
$$
p = p_1 \ NEXT \ p_2 = \{ [s_1 \to s_2] | s_1 \in  p_1, \ \  s_2 \in  p_2 \}
$$  
Note, $s_1$ is one sequence, so is $s_2$. $[s_1 \to s_2]$ is a new sequence that is formed by first $s_1$, then $s_2$. We denote the operator NEXT as $p = p_1 \to p_2$. 
\end{definition}
That is to say, we can connect 2 patterns to form a new pattern. This operator "NEXT" is a very fundamental operation in patterns.

One special case is to connect several spatial patterns to a temporal pattern. If $p_1, p_2, \ldots, p_k$ are a group of spatial patterns, we can connect them to form a temporal pattern $t$ as:
$$
t =  p_1 \to p_2 \to \ldots \to p_k
$$
It is interesting to see an instance of $t$, which is:
$$
s = [b_1 \to b_2 \to \ldots \to b_k], \ \ b_j \in p_j, \ j = 1, \ldots, k
$$

NEXT operator is fundamentally difference from OR, AND.

%% file: spattern.tex
Objective spatial pattern is a subset of base pattern space, objective temporal pattern is a subset of base pattern column. They are independent from any learning machine. But we want to know how a learning machine to perceive objective patterns, which is subjective. So, we have following considerations. When a base pattern or a sequence of base patterns is presented to learning machine $M$, $M$ will generate output. Thus, inside $M$, there must be some place that reflects the input so that the machine can produce correct output. Such places are very crucial.

\begin{definition}[\bf Perception Bit]
Inside a learning machine $M$, if there is one bit $pb$ with such behavior: there is one subset of base pattern column, $S \subset PB^0_N*$, so that for any $s \in S$, the bit has value 1, $pb = 1$, and for any $s \notin S$, the bit has value 0, $pb = 0$, we then call such bit $pb$ as processing bit/perception bit.
\end{definition}

Perception bits play very crucial role in learning machine. For perception bit, we have following simple yet crucial theorem.

\begin{theorem}[\bf Existence of Perception Bits]
For any learning machine $M$, the perception set is not empty.  
\end{theorem}
{\bf Proof: } We first exclude the extreme situation that $M$ always output 0 or always output 1. After excluding the extreme situation, we know for some $p \in PS^0_N*$, output is 0, for some, output is 1. We define all $p \in PS^0_N*$ that output is 1 as one set $B$, called as supporting set. So we can say that the supporting set $B$ is a proper subset of base pattern column $PS_N^0*$, so does  $B^c$. Thus, there must exist one bit $pb$ inside $M$ with such property: for $p \in B$, $pb$ = 1, if $p \notin B$, $pb$ = 0. Thus, $pb$ is a perception bit.

Using perception bits, we can describe perception of a learning machine. Below, we will discuss subjective spatial pattern first, then, subjective temporal pattern,  then, subjective pattern (spatial and temporal).

\subsection{Subjective Spatial Patterns}

We discussed this part before, so we repeat quickly. We start from how to perceive a base pattern.

\begin{definition}[\bf Basic Pattern Perceived by]
$M$ is a learning machine, $\{pb_j |\ j = 1, \ldots, L\}$ are set of perception bits. For 2 basic patterns $b_1$ and $b_2$, the perception bits take values accordingly: $(pv_1^1, pv_2^1, \ldots, pv_L^1)$ and $(pv_1^2, pv_2^2, \ldots, pv_L^2)$, and at least for some $k$, $1 \le k \le L$, $pv_k^1 = pv_k^2 = 1$, we then say, at perception bit $pb_k$, $M$ subjectively perceive $b_1$ and $b_2$ as same.
\end{definition}

Further, we have:

\begin{definition}[\bf Spatial Pattern Perceived by]
$M$ is a learning machine, $\{pb_j |\ j = 1, \ldots, L\}$ are set of perception bits. Suppose $p$ is a set of basic patterns, and at perception bit $pb_k$, $1 \le k \le L$, each basic pattern of $p$ will take value 1, we then say, at $pb_k$, $M$ subjectively perceive $p$ as one pattern. We also say $p$ is a subjective spatial pattern perceived by $M$ at perception bit $pb_k$.
\end{definition}

This definition tells us that a subjective spatial pattern is a set of base patterns, i.e. an objective spatial pattern. But, these base patterns must have one common property: they are perceived same by learning machine at some perception bit. This is very fundamental understanding for subjective spatial pattern. 

Subjective spatial pattern depends on perception of learning machine, so does subjective operators. Here we define subjective operators for subjective spatial patterns.

\begin{definition}[\bf Negation Operator for Subjective Spatial Pattern]
Suppose $M$ is a learning machine, $\{pb_j \ |\ j = 1, \ldots, L\}$ are perception bits. For a subjective spatial pattern $p$ perceived at $pb_k$ by $M$, $q$ is another spatial pattern perceived at $pb_k$ by $M$, but at perception bit $pb_k$, their perception values are all opposite. 
\end{definition}
We denote this pattern $q$ as $q$ = NOT $p$ or $q = \neg p$.

\begin{definition}[\bf OR Operator for Subjective Spatial Pattern]
Suppose $M$ is a learning machine, $\{pb_j \ |\ j = 1, \ldots, L\}$ are perception bits. For any 2 subjective spatial patterns $p_1$ and $p_2$, $p_1$ perceived at $pb_{k_1}$ by $M$, and $p_2$ perceived at $pb_{k_2}$ by $M$, $p$ is another subjective pattern, and perceived by $M$ in this way: first, if necessary $M$ will modify its perception bits so that there is a perception bit $pb_l$, then $M$ perceive any base patterns from either $p_1$ or $p_2$ at perception bit $pb_l$ same.
\end{definition}
This definition tells, if $pb_l$ does not exist, $M$ will generate this perception bit first, and make $p_1$ and $p_2$ perceived same there. We can also say, new pattern $p$ is either $p_1$ or $p_2$ appears. We can denote this new pattern as $p = p_1 \ OR \ p_2$ = $p_1 +_s p_2$. Note, if we want to do operation OR, we might need to modify perception bits of $M$. This is often done by adding a new perception bit. Without the perception bit ready, operation could not be done.

\begin{definition}[\bf AND Operator for Subjective Spatial Pattern]
Suppose $M$ is a learning machine, $\{pb_j \ |\ j = 1, \ldots, L\}$ are perception bits. If  $p_1$ is one subjective spatial pattern perceived at $pb_{k_1}$, $p_2$  is one subjective spatial pattern perceived at $pb_{k_2}$, then, all base patterns that $M$ perceives at both $pb_{k_1}$ and $pb_{k-2}$ at the same time will form another subjective spatial pattern $p$, and $p$ is perceived by $M$ at $pb_l$. 
\end{definition}
This definition tells, if $pb_l$ does not exist, $M$ will generate this perception bit first. We can also say, new pattern $p$ is both $p_1$ and $p_2$ appear together. We can denote this pattern $p$ as $p = b_1 \ AND \ b_2$ = $b_1 \cdot b_2$. Note, if we want to do AND operator, we have to modify perception bits of $M$.

We have 3 subjective operators for subjective spatial patterns. If starting from some base patterns and applying the 3 subjective operators consecutively, we will have one algebraic expression. In \cite{paper2}, we called such algebraic expression as X-form. However, since we are dealing with spatial and temporal patterns, we adapt name to sX-form, "s" stands for spatial. 

\begin{definition}[\bf sX-Form]
If $E$ is one algebraic expression of 3 subjective operators, "$\neg, +, \cdot$", and $g = \{b_1, b_2, \ldots, b_K\}$ is a group of base patterns, then we call the expression $E(g) = E(b_1, b_2, \ldots, b_K)$ as an sX-form upon $g$, or simply sX-form. We note, in order to have this expression make sense, quite often, learning machine $M$ needs to modify its perception bits accordingly. We also call the group of base patterns $g = \{b_1, b_2, \ldots, b_K\}$ as footing of $E$.
\end{definition}

$E$ is an algebraic expression. In a pure sense, it is not connected to anything of learning machine or patterns. It is just one algebraic expression. However, in order this algebraic expression to connect to pattern,  learning machine needs to modify its perception bits so that this expression to make sense, and it must be based on footing. Such situation creates good tool for us so that we can use sX-form to express subjective spatial pattern. Following theorem tells us that in fact sX-form can express any subjective spatial pattern. 

\begin{theorem}[\bf Expressibility of sX-form]
Any objective spatial pattern $p$ can be expressed by one sX-form with footing that has less than $N$ base patterns.  
\end{theorem}
The proof can be seen in \cite{paper2}. We then turn to subjective temporal patterns.

\subsection{Subjective Temporal Patterns}
Spatial pattern only involves one input, but temporal pattern must involve sequence. How sequence is perceived? And, how to turn sequence to subjective temporal pattern?. How subjective spatial pattern is perceived could give us a guide: we first consider how one base pattern is perceived, then consider how a group of base patterns is perceived. So, we need to first consider how to perceive a sequence, then a group of sequences.

\begin{definition}[\bf Sequence Perceived by]
Suppose $M$ is a learning machine, $\{pb_j \ |\ j = 1, \ldots, L\}$ are perception bits. If $[b_1 \to b_2 \to \ldots \to b_k]$ is a sequence of base patterns with length k, and if there are k perception bits, $pb_{j_1}, pb_{j_2}, \ldots, pb_{j_k}$, so that $b_1$ is perceived at $pb_{j_1}$, $b_2$ is perceived at $pb_{j_2}, \ldots$, and $b_1$ is perceived first, $b_2$ is perceived second, $\ldots$,  $b_k$ is perceived at last, and must in such order, and there is one perceive bit $pb_t$ to reflect such perception, i.e. $pb_t =1$ if the sequence is perceived in such way, otherwise, $pb_t = 0$, we will say sequence $[b_1 \to b_2 \to \ldots \to b_k]$ is perceived by $M$ at $pb_t$ as one sequence, and we say the perception bits $pb_{j_1}, pb_{j_2}, \ldots, pb_{j_k}$ are sensing of the sequence.
\end{definition}

Note, one sequence with length $k$, $[b_1 \to b_2 \to \ldots \to b_k]$, has $k$ base patterns. Each of them has to be perceived as spatial pattern first. But, just spatial perception is not good enough. They must also be perceived in the order: $b_1$ first, $b_2$ second, $\ldots$. Also, $M$ must be able to sense such order, and reflect such a sensing to a perception bit. This is quite different than spatial pattern.
 
The perception bit $pb_t$ is crucial. For spatial pattern, there is no such a perception bit. This perception bit is actually reflect the ability of $M$: to be able to sense a sequence and to behave according to the sequence. We will call such perception bit $pb_t$ as temporal perception bit (contrast to spatial perception bit). 

\begin{theorem}[\bf Existence of Temporal Perception Bit]
Suppose $M$ is a learning machine, and suppose $M$ can process at least one sequence, then among its perception bits: $\{pb_j \ |\ j = 1, \ldots, L\}$, temporal perception bits exist. 
\end{theorem} 
{\bf Proof: } Suppose $M$ can process this sequence $s$, and suppose when $s$ inputs, $M$ output 1. There must have some other sequence $s'$, when $s'$ inputs, $M$ output 0. Thus, there exists one bit $pb$ inside $M$ with such property:  for some sequence, $pb$ = 1, for other sequence, $pb$ = 0. So, this $pb$ is one temporal perception bit.

By using temporal perception bit, we can define subjective temporal pattern.

\begin{definition}[\bf Temporal Pattern with Same Length Perceived by]
Suppose $M$ is a learning machine, $\{pb_j \ |\ j = 1, \ldots, L\}$ are perception bits. If $p$ is set of sequences of base patterns with same sequence length, and $pb_t$ is a temporal perception bit, and for all instance of $p$, $s \in p$ ($s$ is a sequence, like $[b_1 \to b_2 \to \ldots \to b_k]$), $s$ is perceived by $M$ at $pb_t$ as one sequence, we then say $p$ is subjectively perceived as one temporal pattern with same length by $M$ at $pb_t$. 
\end{definition}

This definition tells us that a subjective temporal pattern with same length is actually a set of sequence of base pattern, i.e. one objective pattern, however, they are common in such way: they all have the same length, and are perceived as sequence by $M$ at a particular temporal perception bit $pb_t$.

Now, we turn to operators of temporal patterns. First, we consider the operator to connect 2 subjective spatial patterns. Suppose $p_1, p_2$ are 2 subjective spatial patterns, we can connect them to form a subjective temporal pattern by such a way: Suppose $p_1$ is perceived at $pb_{j_1}$, and $p_2$ is perceived at $pb_{j_2}$. Learning machine $M$ might have no sense about the order of perception of $p_1$ and $p_2$. However, if $M$ can adapt some temporal sense, i.e. to perceives $p_1$ first and $p_2$ second, and has one temporal perception bit $pb_t$ to reflect this perceived sequence, $M$ perceives the set $\{ [b_1 \to b_2] \ | b_1 \ is\ an\ instance\ of\ p_1,b_2\ is\ an\ instance\ of\ p_2\}$ as a subjective temporal pattern. 

Very important to notice: in order to make 2 subjective spatial patterns connected to become a subjective temporal pattern, $M$ must do crucial adaption: 1) modify its sense of ordering, 2) if necessary, add one temporal perception bit, 3) use the temporal perception bit to reflect the perceived sequence. This is a lot of adaption. But, this is absolutely necessary to perceive a temporal pattern.

\begin{definition}[\bf Subjective Operator NEXT for Spatial Patterns]
Suppose $M$ is a learning machine, $\{pb_j \ |\ j = 1, \ldots, L\}$ are perception bits. If there are subjective spatial pattern $p_1$ and $p_2$, then a new subjective temporal pattern $t$ can be constructed as:
any instance of $t$ is such a form, $[b_1 \to b_2]$, where $b_1$ is an instance of $p_1$, and $b_2$ is an instance of $b_2$, $M$ perceives $b_1$ first and $b_2$ second, and has one temporal perception bit $pb_t$ to reflect this sequence. We call this operator NEXT, and denote  $t = e_1 \to e_2$. 
\end{definition}

We call $p_2$ is next to $p_1$, and we say that 2 spatial patterns are connected to form a temporal pattern. Of course, in order to make such operator work, it absolutely necessary to make a lot of adaption described above. Compare to necessary adaption for spatial pattern operator, the adaption for temporal pattern operator is much more. This is understandable and is crucial.

We can consecutively apply the NEXT operator to a group of spatial patterns $p_1, p_2, \ldots, p_k$, we will have: $t = p_1 \to p_2 \to \ldots \to p_k$. This is a subjective temporal pattern generated from  $p_1, p_2, \ldots, p_k$. Of course, in order to make sense, a lot of adaption are necessary. 

In short, the above subjective temporal pattern is generated by a group of subjective spatial patterns. But, we also know subjective spatial pattern can be expressed by sX-form, it would be better to define NEXT operator to sX-form.

\begin{definition}[\bf Subjective Operator NEXT for sX-form]
Suppose $M$ is a learning machine, $\{pb_j \ |\ j = 1, \ldots, L\}$ are perception bits. We know any sX-form represent a subjective spatial pattern. For any 2 sX-form $e_1$ and $e_2$, we can generate a new temporal pattern $t = p_1 \to p_2$, where $p_1$ and $p_2$ are the subjective spatial patterns represented by $e_1$ and $e_2$ accordingly. Subjective operator NEXT for the 2 sX-forms is the operator NEXT for spatial patterns $p_1$ and $p_2$. 
\end{definition}

We denote this operator as: $t = e_1 \to e_2$. Of course, in order to make sense for this expression, we have to adapt $M$ a lot, as discussed above. 

We can also consecutively apply this operator, i.e. $e_1 \to e_2 \to \ldots \to e_k$. We give a definition for this kind of expression. 

\begin{definition}[\bf tX-Form]
If $e_1, e_2, \ldots, e_k$ are a group of sX-form, then we call expression $e_1 \to e_2 \to \ldots \to e_k$ as a tX-form.
\end{definition}
 
That is to say, sX-form connected by operator "$\to$" form a tX-form. Here "t" stands for temporal. This form $t$ in fact expresses a subjective temporal pattern. We also say, new pattern $t$ is formed by $e_2$ next to $e_1$. We can denote the operator as "$\to$", $t = e_1 \to e_2$.

\begin{theorem}[\bf Expressibility of tX-form]
Any objective temporal pattern with same length $p$ can be expressed by one tX-form.  
\end{theorem}
{\bf Proof:} Suppose $p = \{s_1, s_2, \ldots\}$, each $s_j$ is an instance of $p$ and is a sequence with length $k$. So, $s_j = [b_{j_1} \to b_{j_2} \to b_{j_k}]$. Now, we know this set $\{b_{j_i} |\ i = 1, 2, \ldots$ is a set of base patterns, according to Theorem 3.2, it can be expressed by one sX-form $p_i$. We can see clearly that $p = p_1 \to p_2 \to \ldots \to p_k$. 

This theorem tells us that tX-form is good enough to express any objective temporal pattern with same length.

\subsection{Subjective Patterns}
From subjective spatial patterns, we derive sX-form, which can express any spatial pattern. From subjective temporal patterns with same length, we derive tX-form, which can express any temporal pattern with same length. What about any subjective pattern, spatial and temporal?


Let's consider some examples first. If $b_1, b_2$ are 2 base patterns. Then, $p = b_1 + b_2$ is a sX-form expresses a subjective spatial pattern, $[b_1 \to p]$ is a tX-form, expresses a subjective temporal pattern with length 2. But, how about one subjective pattern that "either $[b_1 \to p]$ or $b_2$"? This is surely one legitimate subjective pattern, but it is not purely spatial pattern, nor purely temporal pattern. How a machine to perceive it? How to express this pattern algebraically? 

\begin{definition}[\bf Subjective Pattern Perceived by]
Suppose $M$ is a learning machine, $\{pb_j \ |\ j = 1, \ldots, L\}$ are perception bits. If $p$ is subset of base pattern column, i.e. $p \subset PS^0_N*$, if there is one perception bit $pb$ so that any instance of $p$ is perceived same by $M$ at $pb$, we then say $p$ is subjectively pattern perceived by $M$ at $pb$. 
\end{definition}   

We can define 2 subjective operators as below. They are defined before, but, we make some modification so that it can be applied to any subjective pattern.

\begin{definition}[\bf OR Operator for Subjective Pattern]
Suppose $M$ is a learning machine, $\{pb_j \ |\ j = 1, \ldots, L\}$ are perception bits. For any 2 subjective patterns $p_1$ and $p_2$, $p_1$ perceived at $pb_{k_1}$ by $M$, and $p_2$ perceived at $pb_{k_2}$ by $M$, $p$ is another subjective pattern, and perceived by $M$ in this way: first, if necessary $M$ will modify its perception bits so that there is a perception bit $pb_l$, then $M$ perceive any instance of either $p_1$ or $p_2$ same at perception bit $pb_l$.
\end{definition}
This definition tells, if $pb_l$ does not exist, $M$ will generate this perception bit first, and make $p_1$ and $p_2$ perceived same there. We can also say, new pattern $p$ is either $p_1$ or $p_2$ appears. We can denote this new pattern as $p = p_1 \ OR \ p_2$ = $p_1 +_s p_2$. Note, if we want to do operation OR, we might need to modify perception bits of $M$. This is often done by adding a new perception bit. Without the perception bit ready, operation could not be done.

\begin{definition}[\bf NEXT Operator for Subjective Pattern]
Suppose $M$ is a learning machine, $\{pb_j \ |\ j = 1, \ldots, L\}$ are perception bits. If there are subjective pattern $p_1$ and $p_2$, then a new subjective pattern $t$ can be constructed as:
any instance of $t$ is such a form, $[b_1 \to b_2]$, where $b_1$ is an instance of $p_1$, and $b_2$ is an instance of $b_2$, $M$ perceives $b_1$ first and $b_2$ second, and has one temporal perception bit $pb_t$ to reflect this sequence. We call this operator NEXT, and denote  $t = e_1 \to e_2$. 
\end{definition}

Starting from some base patterns, and apply the operators we defined ("+, $\cdot$, $\neg$" for subjective spatial patterns, "+, $\to$" to any subjective patterns), we can have X-form.  

\begin{definition}[\bf X-form as Algebraic Expression]
If $E$ is one algebraic expression of  5 subjective operators, "$\neg, +, \cdot$" for subjective spatial patterns, and "$+, \to$" for any other subjective patterns, $g = \{b_1, b_2, \ldots, b_K\}$ is a group of base patterns, then we call the expression $E(g) = E(b_1, b_2, \ldots, b_K)$ as an X-form upon $g$, or simply X-form. We note, in order to have this expression make sense, quite often, learning machine $M$ needs to modify its perception bits accordingly. We also call the group of base patterns $g = \{b_1, b_2, \ldots, b_K\}$ as footing of $E$.
\end{definition}

So, can X-form good enough to express any subjective pattern? Yes, it can.

\begin{theorem}[\bf Expressibility of X-form]
Any objective pattern $p$ (i.e. any subset of base pattern column $PS^0_N*$) can be expressed by one X-form with footing that has less than $N$ base patterns.  
\end{theorem}
{\bf Proof:}  We can group $p$ by sequence length. For so that $p = p_1 + p_2 + \ldots$, each $p_l$ is set of sequence with length $l$. According to Theorem 3.4, $p_l$ can be expressed by a tX-form. We can also make the footing for all tX-form as a same number of base patterns, and this number is less than $N$.

This theorem tells us any subjective patterns can be expressed by one X-form with footing that has less than $N$ base patterns. Shortly say, X-form has strong expressibility for all subjective patterns.

%% file: xform.tex
In last section, we discussed subjective spatial patterns and sX-form, subjective temporal patterns and tX-form, and subjective patterns and X-form. X-form is the fundamental tool for us. In definition 3.16, X-form is one algebraic expression. However, when this expression is combined with footing, it is a subjective pattern. 

\begin{definition}[X-form as Subjective Pattern]
If $E = E(g) = E(b_1, b_2, \ldots, b_K)$ is X-form, then it is an algebraic expression of  5 subjective operators ("$\neg, +, \cdot$" for subjective spatial patterns, and "$+, \to$" for any other subjective patterns). The $K$ base patterns $b_1, b_2, \ldots, b_K$ is called footing. Combining with footing (if make sense), $E$ actually expresses a subjective patterns that is generated from $b_1, b_2, \ldots, b_K$ with those operations.
\end{definition}

That is to say, for an algebraic expression $E$, if it combines footing $\{b_1, b_2, \ldots, b_K\}$, if make sense, then we have a subjective pattern $E(b_1, b_2, \ldots, b_K)$. This gives us one powerful tool to deal with subjective pattern. However, X-form can represent information processing as well.

\begin{definition}[X-form as Information Processing]
If $E = E(g) = E(b_1, b_2, \ldots, b_K)$ is X-form with footing $b_1, b_2, \ldots, b_K$. if it makes sense, $E$ expresses a subjective pattern $p$, further, it is an information processing by such a way: for any sequence $s \in PS^0_N*$, if $s \in p$, $E(s) = 1$, if $s \notin p$, $E(s) = 0$.
\end{definition}

So, X-form is one algebraic expression, X-form represents a subjective pattern (spatial and temporal), X-form is an information processing. Actually, X-form is all of them. Such an multi-role gives us a very powerful tool to deal with patterns and learning machine. 

\begin{example}[\bf  Simple Examples] Here, $b_1, b_2, \ldots$ are base patterns. \\
1. $e_1 = b_1 + b_2$ is one X-form. Actually, it is one sX-form. \\
2. $e_2 = b_1 + b_2 \cdot b_2$ is one X-form. It is also sX-form.\\
3. $e_3 = (b_1 + b_2) \cdot (b_1 + b_3)$ is one sX-form. \\
4. $e_4 = [b_1 \to e_2 \to e_3]$ is one tX-form. \\
5. $e_5 = e_4 + ([e_1 \to e_4])$ is one X-form.  
\end{example}

For subjective spatial operators, we also have:

{\bf  Subjective spatial operators:}\\
Commutative: $a + b = b + a$，$a \cdot b = b \cdot a$ \\
Associative: $a + (b + c) = (a + b) +c, a \cdot (b \cdot c) = (a \cdot b) \cdot c$ \\
Distribution: $a \cdot (b + c) = (a \cdot b) + (a \cdot c)$ \\
Negation reflect: $\neg (\neg a) = a$ 

{\bf  Subjective operators:}\\
Commutative: $a + b = b + a$ \\
Associative: $a + (b + c) = (a + b) +c$ \\
Distribution with NEXT: $a \to (b + c) = (a \to b) + (a \to c), (a +b) \to c = (a \to c) + (b \to c)$ 

However, in order to make sense, perception bits must be adapted. There are more properties of these operators. \\

Here, we make some comments for X-form.

{\bf Multiple Role:} \\
The multiple roles played by X-form is very useful. First, as an algebraic expression, algebraic operation can be easily conducted. Then, the algebraic expression can combine with base patterns to express a subjective pattern. And, X-form is also an information processing, which is not opaque to us, on the contrary, it is a clear logical statement to tell how learning machine is processing information. These are perfect for us to study learning machine.

We can compare X-form with current deep learning, which is quite opaque, and no way to do algebraic operation on perception. We can also compare X-form with category theory, which is used to describe the dynamics of concepts. While category theory allows explicit algebraic operations on concepts, it is very difficult to link the concepts to learning machine.  
X-form by its definition is logical statement, and the other hand, it is connectionist by nature. It combines connectionist approach and symbolic approach naturally.

X-form itself is a logical relationship (if the X-form is very long, the logic may be complicated), so X-form can be naturally associated with various logical reasoning within the learning machine. These logical reasonings do not necessarily have to be classic. They can be non-classical or probabilistic. The work in these areas has yet to be expanded. In short, the X-form gives us a great tool for subjective view of a learning machine, and it is a mathematical object that is well worth to study in depth.


{\bf Relationship of Spatial and Temporal:} \\
Spatial means to consider information in data arrangement in base pattern. Quite differently, temporal involve sequence. More precisely, a sequence has to input a base pattern first, then another base pattern, etc. Learning machine will use not only information cross index, but also use information contains in ordering, and previous and afterwards input. Question is: can spatial pattern turns to be temporal, and vise versa? Can we convert temporal pattern to spatial pattern with a bigger dimension? On the other hand, can we convert spatial pattern to temporal pattern by cutting base pattern space into smaller pieces and make them to be perceived in order? Actually, we can. We can turn spatial to temporal, and vise versa. In fact, to do so, not only is possible, but also very desirable. In many situation, we need to do so. Converting spatial and temporal gives us a very powerful tool. 

The great thing is: X-form can capture converting spatial and temporal easily. Algebraically, X-form can express all of them clearly and easily.  

Finally, due to the role X-forms play in learning machine, we would like to formulate one conjecture as below:

\begin{conjj}[\bf X-form]
Any computing system, if it has some universal learning capability for spatial and temporal patterns,  inside it there must have some structure equivalent to sX-form, tX-form and X-form. 
\end{conjj}

%% file: more.tex
Inside a learning machine, there are many X-forms to express subjective view of learning machine. X-form can express well those subjective views. This is a very important step towards universal learning machine. Yet, it is only the first step. There are a lot of research ahead. We need to understand how to modify the subjective view of a learning machine so that this view can match the environment and data. This process to match is actually the learning. In fact, from this point of view, learning is the dynamics of X-form in conceiving space. The dynamics of X-form will be our future work. Actually, we did some parts in \cite{paper2, cpaper}, but only for sX-form. We have not doubt that this will be one very rich research area. 

Actually, many authors think inside brain of animal and human, just as inside a learning machine, there are a lot of "pattern recognizer" (for example \cite{rkur}), and these "pattern recognizer" play crucial role in perception and action. We think the X-form as one algebraic expression for "pattern recognizer" could help us on further research in these areas.